\documentclass[conference]{IEEEtran}
\IEEEoverridecommandlockouts
\usepackage{cite}
\usepackage{amsmath,amssymb,amsfonts}
\usepackage{algorithmic}
\usepackage{graphicx}
\usepackage{textcomp}
\usepackage{xcolor}
\usepackage{algorithm}
\usepackage{subfig}
\usepackage{url}
\def\BibTeX{{\rm B\kern-.05em{\sc i\kern-.025em b}\kern-.08em
    T\kern-.1667em\lower.7ex\hbox{E}\kern-.125emX}}
\begin{document}

\title{
Camera Perspective Transformation to Bird's Eye View via Spatial Transformer Model for Road Intersection Monitoring
}

\author{\IEEEauthorblockN{Rukesh Prajapati, and Amr S. El-Wakeel}
\IEEEauthorblockA{\textit{Lane Department of Computer Science and Electrical Engineering} \\
\textit{West Virginia University}\\
Morgantown, USA \\
rp00052@mix.wvu.edu, amr.elwakeel@mail.wvu.edu}
}

\maketitle

\begin{abstract}
Road intersection monitoring and control research often utilize bird’s eye view (BEV) simulators. In real traffic settings, achieving a BEV akin to that in a simulator necessitates the deployment of drones or specific sensor mounting, which is neither feasible nor practical. Consequently, traffic intersection management remains confined to simulation environments given these constraints. In this paper, we address the gap between simulated environments and real-world implementation by introducing a novel deep-learning model that converts a single camera’s perspective of a road intersection into a BEV. We created a simulation environment that closely resembles a real-world traffic junction. The proposed model transforms the vehicles into BEV images, facilitating road intersection monitoring and control model processing. Inspired by image transformation techniques, we propose a Spatial-Transformer Double Decoder-UNet (SDD-UNet) model that aims to eliminate the transformed image distortions. In addition, the model accurately estimates vehicles positions, and enables the direct application of simulation-trained models in real-world contexts. SDD-UNet model achieves an average dice similarity coefficient (DSC) above 95\% which is 40\% better than the original UNet model. The mean absolute error (MAE) is 0.102 and the centroid of the predicted mask is 0.14 meters displaced, on average, indicating high accuracy.

\end{abstract}

\begin{IEEEkeywords}
Road intersection monitoring, BEV, perspective transformation, spatial transformer, double decoder UNet.
\end{IEEEkeywords}

\section{Introduction}

Intelligent Transportation Systems (ITS) play a critical role in advancing traffic management and safety by utilizing various and complementary technologies. Effective coordination of road intersections is essential for optimizing traffic flow, reducing congestion, and minimizing accident risks, thereby ensuring efficient and secure transportation networks. Recent research has extensively focused on optimizing traffic management at road intersections. Advanced deep-learning models have been developed and trained to control traffic lights based on the vehicle count approaching the intersection\cite{icais1}. These approaches have demonstrated the potential to enhance efficiency by minimizing vehicle wait times \cite{liu1}. However, collecting real-world data for training these models poses significant challenges, as the real-world implementation of training scenarios could lead to accidents. When vehicles are deployed in real-world scenarios for training and data collection using deep reinforcement learning, the multitude of possible actions and decisions can potentially result in collisions. \cite{iv}.  Consequently, the development and training of such models for junction monitoring and control would be more safe and dynamic if conducted using road and traffic simulators.

In literature, researchers utilized road and traffic simulators that offer a bird's eye view (BEV) or aerial perspective of road intersections \cite{icais1, liu1}. The BEV is particularly effective for estimating the position and state of a specific region or the surrounding environment of a given agent or vehicle \cite{Reiher2020ASD}. In literature, Simulation of Urban MObility (SUMO), an open-source traffic simulation software was commonly adopted as it enables the construction of road networks, integration of automated vehicles, combining different modes of transportation, etc. \cite{sumo}. However, a significant challenge of road and traffic network simulators like SUMO is their exclusive reliance on BEV for scenario representation. In real-world settings, continuously obtaining a BEV would require a drone to be stationed over the intersection or specific sensor mount which is not always practical and efficient \cite{highDdataset}. Therefore, for camera-based-intersection monitoring, there is a need for the capability of generating a BEV of a scenario using camera/s installed at the intersection, that can be mounted on light/utility poles, traffic lights, roadside units, or nearby infrastructure.

Recent research on perception utilizing BEV has focused on vehicle-level perception\cite{cvpr1, cvpr2, cvpr3}. This approach aids in collision avoidance and navigation by providing comprehensive information about the surrounding environment. Prominent deep learning models typically employ multi-camera feeds to generate BEV images\cite{eccv1}. A similar system could be adapted for traffic monitoring at intersections. However, establishing a multi-camera setup poses challenges, including the need for synchronization, camera connectivity to the system, and ensuring coverage of all road segments converging at the intersection \cite{multi_cam}.

The conversion of road intersection views to BEV, was recently explored  \cite{lidar1}. Approaches involving the fusion of LiDAR and camera data, or using a standalone LiDAR, have been considered \cite{2023arXiv230301212S}. However, LiDAR has several challenges as it operates using point clouds and does not detect objects as efficiently as cameras for long distances. As objects move farther from the LiDAR sensor, the distance between laser beams increases due to resolution constraints, potentially causing objects within range to go undetected\cite{lidar2}. Additionally, LiDAR systems are less affordable compared to vision cameras. Therefore, cameras can outperform LiDAR for intersection monitoring under clear weather conditions.

A perspective transformation of an image capturing the entire intersection can be utilized to create a BEV of a road intersection. However, this requires first calculating the transformation matrix, which involves placing reference objects within the scene and computing the matrix using known points from the perspective view and the corresponding BEV \cite{Reiher2020ASD}. This method is neither time-efficient nor practical. Additionally, the transformed image often suffers from distortion due to the interpolation technique used during the transformation process, complicating object detection and position estimation. Therefore, a more efficient perspective transformation system is needed to address these limitations when using a single camera. Spatial transformers focus on spatial manipulation of data, while self-attention transformers emphasize contextual relationships within data. As we focus on the pole camera's spatial transformation, we utilize the spatial transformer.

In this paper, we propose a novel deep-learning model inspired by image transformation techniques to convert the perspective of a single camera mounted to monitor a road intersection into a BEV. While most of the traffic management research is conducted in simulation environments using BEV, our work bridges the gap between these simulations and real-world applications,  by accurately transforming the vehicle's position in BEV. BEV generated by our model includes only the shape and position of the vehicle as seen from a top-down perspective. This representation does not capture any of the features present on the vehicle in a real-life scenario. Consequently, the output can be regarded as a binary mask that can be superimposed onto the top-view scene. For data collection, we developed a 3D simulation environment that replicates real-world scenarios to ensure the model's applicability in actual traffic junctions. This simulated environment facilitates data acquisition without the need for a real-world setup, thereby reducing the complexity associated with deploying drones for BEV ground truth generation and data labeling. The model processes camera perspective images from traffic intersections and generates a BEV representation of vehicle positions. This BEV output can then be utilized as input for models trained using BEV simulators. Thus, the proposed system and model utilizes a single camera to capture the entire junction, thereby avoiding the complexities associated with multi-sensor setups, coordination, or fusion.

\section{Framework and Methodology}
\subsection{Simulation Environment and Dataset}

\begin{figure}[t!]
\centerline{\includegraphics[width=65mm,height=3in]{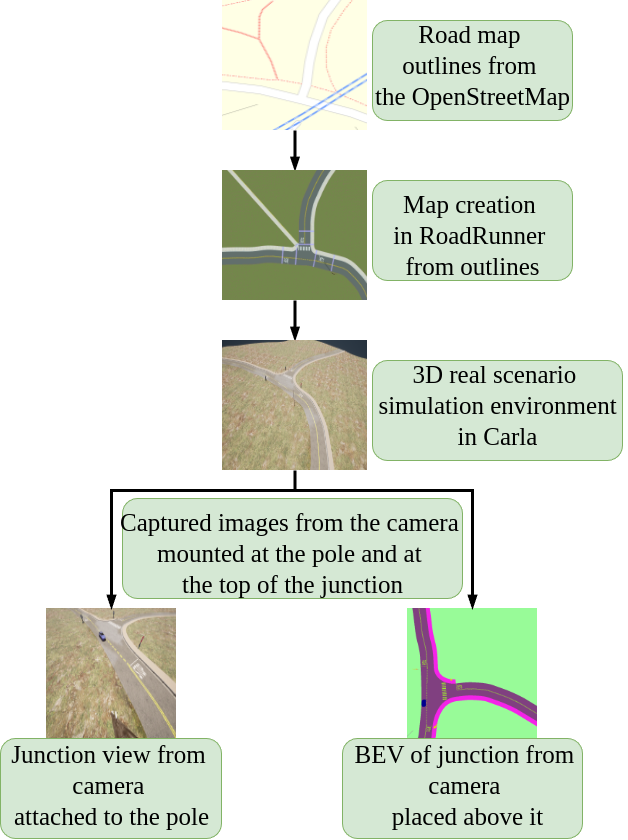}}
\caption{Data collection process}
\label{fig2}
\end{figure}

We developed a simulation environment that closely mimics the real world to facilitate data collection. The detailed process is illustrated in Fig. \ref{fig2}. Initially, we obtained the OpenStreetMap (OSM)\cite{OpenStreetMap} data for the desired location, which provided the outlines of roads and infrastructure. These outlines were imported into RoadRunner software, where a road network was constructed over them to ensure the physical features of the roads matched those in the real world\cite{roadrunner}.

The simulation environment was further enhanced using CARLA \cite{carla}, an open-source software platform that supports the creation of urban layouts, buildings, and vehicles (actors). CARLA allows the integration of various sensors, such as cameras, LiDARs, and radars, for data collection within the simulated environment.

In our simulation setup, the map imported from RoadRunner was used, and two cameras were placed within the environment. The first camera, a standard RGB camera, was mounted on top of a pole near a junction to simulate a real-life surveillance camera monitoring the junction. The second camera, a semantic segmentation camera provided by CARLA, was positioned directly above the junction, facing it.

The semantic segmentation camera images were processed to filter out the background (assigned a value of 0) and retain only the pixels representing vehicles (assigned a value of 1). This processing provided the exact location and shape of the vehicles in the BEV.

\begin{figure}[b!]
\centerline{\includegraphics[width=65mm,height=1.5in]{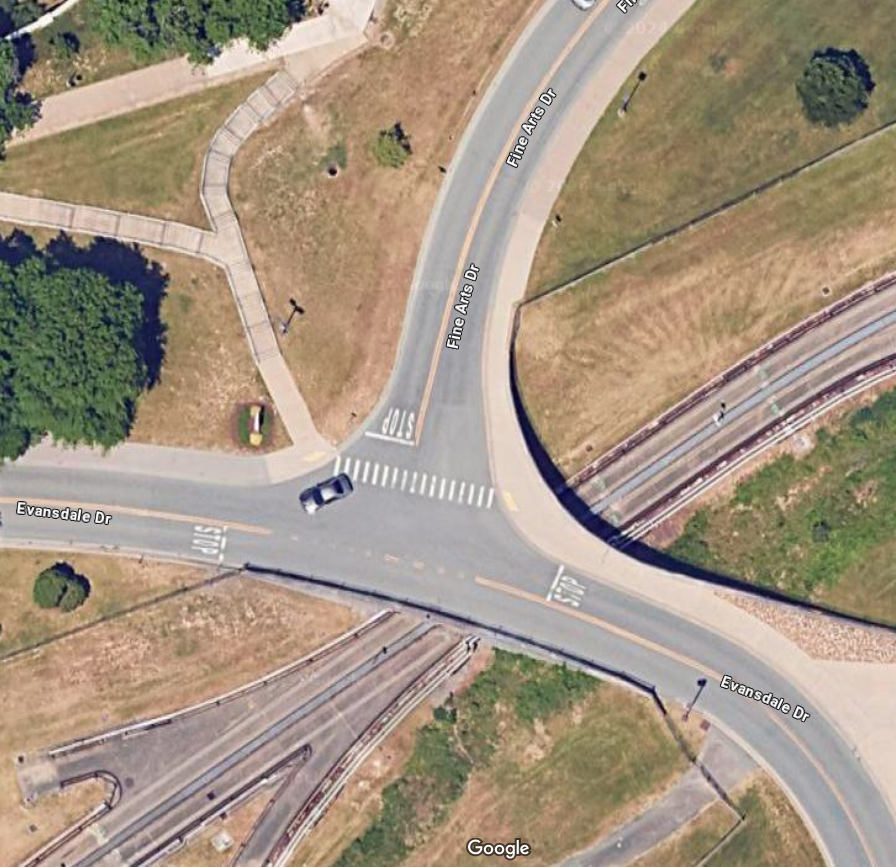}}
\caption{Junction at West Virginia University replicated to create a simulation environment for data collection \cite{googlemaps}}
\label{fig_google}
\end{figure}

\begin{figure*}
\centering
\includegraphics[width=6.5in, height=2.5in]{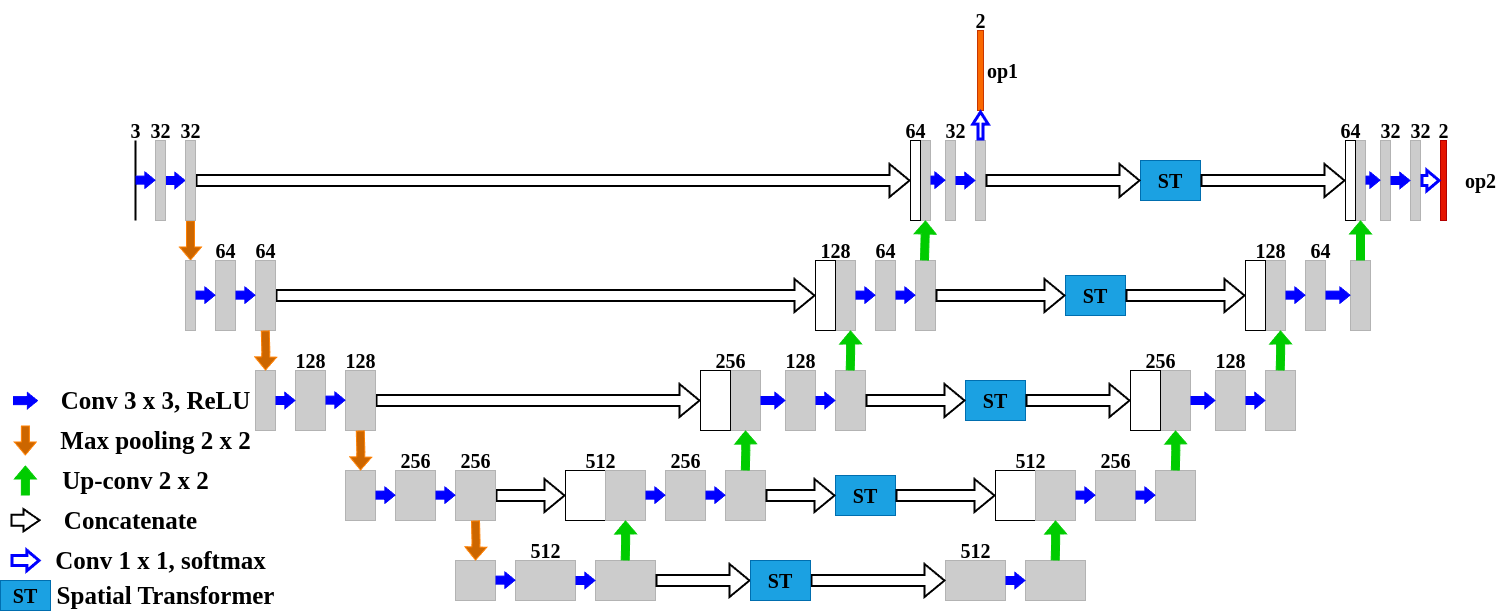}
\caption{Architecture of proposed SDD-UNet model with single encoder and double decoder branch, where the first branch predicts a mask (op1) that comprises a bounding box for the vehicle inside the image and the second branch produces the mask (op2) for the BEV of vehicles position}
\label{fig_model}
\end{figure*}

For our study, we focused on the junction near the Mineral Resources Building at West Virginia University, Morgantown, USA, which is shown in Fig. \ref{fig_google}. The central coordinates of the junction are 39°38'50.9"N, 79°58'30.0"W. Following the previously outlined process, we constructed a simulation environment, subsequently, positioning cameras and vehicles within this setting. For this research, we confined our simulations to scenarios involving a single vehicle. Consequently, only one vehicle was present in the simulation at any given time. We collected 26,578 images and their corresponding BEV from the simulator. The split ratio for the train, validation, and test is 70:10:20.

\subsection{Spatial-Transformer Double Decoder-UNet Model Architecture}
The proposed model Spatial-Transformer Double Decoder-UNet (SDD-UNet) draws inspiration from the UNet architecture, featuring distinct encoder and decoder components. Illustrated in Fig. \ref{fig_model}, the SDD-UNet consists of a single encoder and dual decoder branches. The encoder follows the design principles of the original UNet model, where each pair of convolutional layers is succeeded by a downsampling operation. Each convolutional layer employs a 3x3 kernel with the same padding and a stride of 1. The number of convolutional layers is indicated in the Fig. \ref{fig_model}. For downsampling, a 2x2 max pooling operation is utilized after each convolutional block.

Among the two decoder branches, the first one closely follows the original UNet architecture \cite{unet}. In this branch, each pair of convolutional layers is succeeded by a transpose convolution. The structure of the convolutional layers is similar to that of the encoder. The transpose convolution layers reduce the number of channels by half while doubling the spatial dimensions of the input. The output of each transpose convolution is concatenated with the corresponding encoder features of the same depth, forming a skip connection. The output from the first branch is compared with a mask generated using bounding boxes from the YOLOv8 model\cite{yolo}. This branch is designed to train the model to detect vehicles and determine their positions within the original image.

 The second branch integrates a proposed spatial transformer (ST) in every skip connection before concatenating features from the first decoder branch as shown in Fig. \ref{fig_model}. In addition, the second branch is trained to capture and learn the transformation to BEV. While the first branch focuses on vehicle detection, the second branch leverages the features learned by the first branch, transforming them to simulate the perspective transformation for the detected vehicles.

The dice similarity coefficient (DSC) was employed as the loss function in our model. During the gradient descent optimization process, the goal is to minimize the loss. To utilize DSC as a loss function, we adopted its negative value. Consequently, minimizing this negative DSC effectively maximizes the DSC. The overall loss of the model is calculated as the sum of the DSC from the first branch and the DSC from the second branch.

\subsection{Spatial Transformers}
    The spatial transformers consist of three stages, localization network, grid generator, and grid sampler, as detailed below:
    
\subsubsection{Localization Network}
The localization network of the ST predicts the transformation parameters $\theta$ from the input feature map $U$ \cite{ST}. This network typically consists of convolutional and fully connected layers. Our model comprises two convolutional layers with kernel size of 11x11 and 7x7 with stride of 5 and 3 respectively, and followed by two fully connected layers. The transformation parameters $\theta$ can represent various types of transformations, such as affine transformations. Mathematically, the output of the localization network can be expressed as:
\begin{equation}
    \theta = f_{loc}(U)
\end{equation}
where $f_{loc}$ is a function representing localization network.

\subsubsection{Grid Generator}
The grid generator uses the predicted transformation parameters $\theta$ to create a sampling grid. This grid defines where to sample points in the input feature map to obtain the output feature map. For our perspective transformation, the grid generation can be described as follows:
\begin{equation}
\begin{pmatrix}
x_{s} \\
y_{s}
\end{pmatrix}
=\mathcal{T}_\theta(G)
= \begin{pmatrix}
\theta_{11} & \theta_{12} & \theta_{13} \\
\theta_{21} & \theta_{22} & \theta_{23} \\
\theta_{31} & \theta_{32} & \theta_{33}
\end{pmatrix}
\begin{pmatrix}
x_t \\
y_t \\
1
\end{pmatrix} 
\label{t_matrix}
\end{equation}
Here, \( (x_t, y_t) \) are the target coordinates in the output feature map, and \( (x_s, y_s) \) are the corresponding source coordinates in the input feature map.

\subsubsection{Grid Sampler}
\begin{figure}[b!]
\centerline{\includegraphics[width=70mm,height=1in]{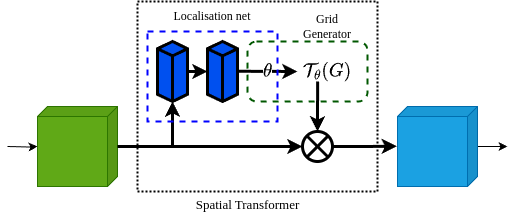}}
\caption{Illustration of Spatial Transformer}
\label{fig_st}
\end{figure}

The sampler uses the sampling grid to produce the output feature map by sampling the input feature map at the grid locations as shown in Fig. \ref{fig_st}. This step typically involves interpolation, such as bilinear interpolation. The sampling process can be mathematically described by:
\begin{equation}
\begin{aligned}
    V(i, j) = \sum_{n=1}^H \sum_{m=1}^W U(m, n) \cdot \max(0, 1 - \\|x_s - m|) \cdot \max(0, 1 - |y_s - n|) 
\end{aligned}
\end{equation} 
where \( V(i, j) \) is the output feature map, \( U(m, n) \) is the input feature map, and \( (x_s, y_s) \) is the source coordinates calculated by the grid generator. This equation represents bilinear interpolation, where the values of the input feature map \(U\) are weighted by their proximity to the sampling points \((x_s, y_s)\).

\subsection{Evaluation Metrics}
We employed the DSC, Mean Absolute Error (MAE), and the pixel distance between the ground truth centroid and the predicted mask centroid as evaluation metrics. 

\begin{equation}
    DSC = \frac{2TP}{2TP+FP+FN}
\end{equation}

\begin{equation}
    MAE = \frac{\left| y - \hat{y}\right|}{n}
\end{equation}

where TP is true positive, FP is false positive, FN is false negative, TN is true negative, $y$ is actual value, and $\hat{y}$ is predicted value.

\section{Results and Discussion}
\subsection{Computer Vision Algorithmic Approach}
An algorithmic approach is used to achieve the perspective transformation that involves the utilization of OpenCV library\cite{opencv_library}. Initially, four reference objects are positioned within the simulation environment. Images are then captured using both a pole-mounted camera and a BEV camera. By identifying corresponding points in the images from both cameras, a transformation matrix can be computed, as described in Eq. \ref{t_matrix}. This matrix enables the transformation of the entire image into the BEV perspective.

\begin{figure}[b!]
\centerline{\includegraphics[width=85mm,height=2.5in]{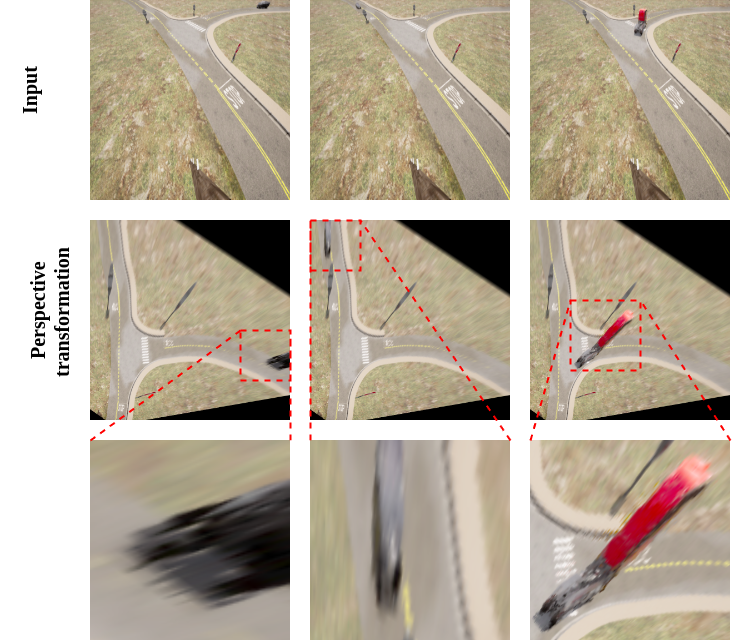}}
\caption{Illustration of distortion in image from perspective transformation}
\label{fig_cv}
\end{figure}

We also computed the transformation matrix for our setup. Fig. \ref{fig_cv} illustrates the outcome of this transformation. The results indicate that the algorithmic transformation is ineffective in this scenario. The transformation introduced significant distortion to the image, resulting in an elongation of certain portions that compromise the visual features. Consequently, neither deep learning models nor the human eye can accurately identify the vehicle in the transformed image. The elongation of objects further complicates the calculation of their centroids, thereby impeding the accurate estimation of vehicle positions. Our model is capable of addressing these issues effectively.

\subsection{Deep Learning Approach}
In this study, we compared three different deep learning models, all developed or modified from the UNet architecture.

\textbf{UNet:} The first model is a standard UNet without any modifications. It comprises an encoder and a decoder. In this model, skip connections are implemented by concatenating layers from the encoder with the corresponding layers in the decoder.

\textbf{UNet(ST skip):} The second model retains the same encoder and decoder structure as the standard UNet. However, the modification lies in the skip connections. Instead of direct concatenation, this model incorporates ST as detailed in Section II(C).

\textbf{SDD-UNet:} The third and proposed SDD-UNet model features a single encoder and two decoder branches. In the first decoder branch, layers are directly concatenated from the encoder. In contrast, the second decoder branch transforms the layers from the first decoder using spatial transformers before concatenation. The detailed structure of this model is provided in Section II(B).

\subsubsection{Training Comparison}
\begin{figure}[b!]
\centerline{\includegraphics[width=70mm,height=2.5in]{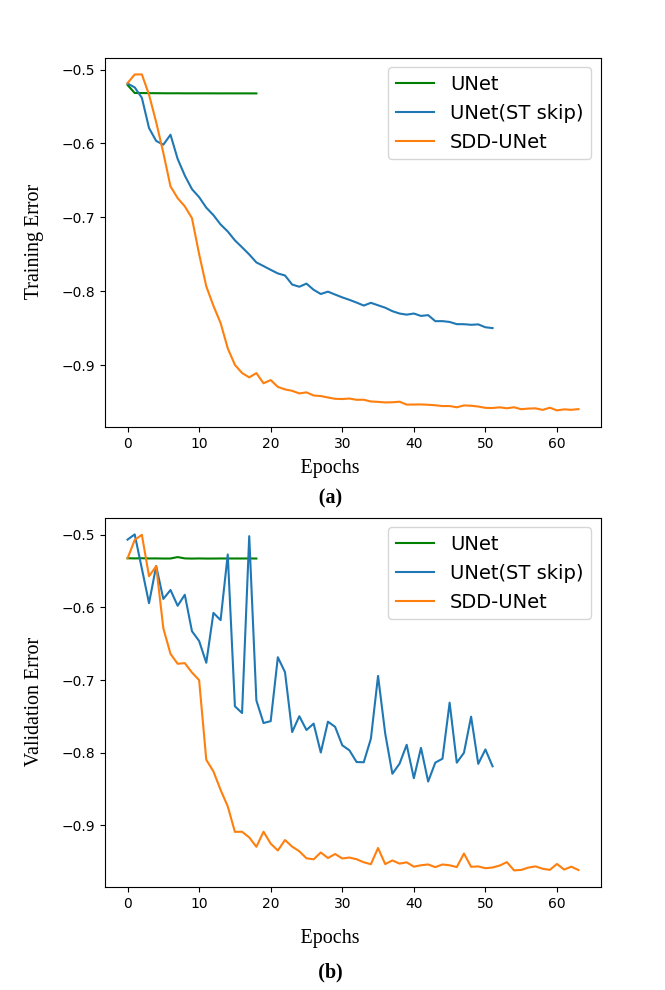}}
\caption{(a) Training error and (b) validation error (Negative DSC is used as an error to update models)}
\label{fig_train}
\end{figure}

We employed early stopping, with a patience of up to 10 epochs, to train all three models. The learning curves are depicted in Fig. \ref{fig_train}. Fig. \ref{fig_train}(a) illustrates the training loss, while Fig. \ref{fig_train}(b) presents the validation loss during the training process.

From the training and validation loss curves shown in Fig. \ref{fig_train}, it is evident that for the standard UNet model, both the validation and training losses did not decrease with additional training epochs. This indicates that the model was unable to learn the transformation necessary to generate the BEV of the vehicles at the intersection. Consequently, the early stopping mechanism terminated the training earlier for this model compared to the others. In contrast, the second model, UNet with ST skip connections, demonstrated some ability to learn the transformation. The DSC loss dropped below -0.8, indicating that the dice similarity during training exceeded 80\%. However, there was no significant improvement beyond this point.

The SDD-UNet outperformed the previous two. Both the training and validation losses fell below -0.90, suggesting that the model achieved more than 90\% dice similarity during training, indicating better learning and transformation capabilities.

\subsubsection{Comprehensive Results}

The test results are presented in Table \ref{tab1}. The original UNet model failed to effectively learn during training, achieving an average DSC of only 0.53 on the test data. Consequently, the model was unable to accurately generate the BEV of the vehicles, resulting in an average centroid distance of 343 pixels from the ground truth centroid. Given that 34.23 pixels approximately correspond to 2.86 meters in the BEV, the average centroid distance for the original UNet was calculated to be 28.74 meters. In contrast, the second model, UNet(ST skip), successfully learned the required transformation, yielding an average DSC of 0.84 on the test data. The centroid distance for this model was 27.02 pixels, equivalent to 2.23 meters.

The SDD-UNet demonstrated superior performance, with an average DSC of 0.957. The centroid distance between the predicted mask and the ground truth was only 1.75 pixels or 0.14 meters. This level of positional accuracy is highly effective for vehicles in the intersection and can be utilized in further processing without the need for additional manual adjustments.

\begin{table}[H]
\caption{Comparison of results for BEV masks generation}
\begin{center}
\begin{tabular}{c|c|c|c}
\hline
\textbf{Models} & \textbf{DSC}& \textbf{MAE}& \textbf{Centroid}\\
 & & & \textbf{Distance (pixel)}\\
\hline
UNet & 0.531$\pm$0.11 & 1.113$\pm$0.57 & 343.99$\pm$188.2\\
UNet(ST skip) & 0.841$\pm$0.135 & 0.385$\pm$0.503 & 27.02$\pm$79.86\\
SDD-UNet & 0.957$\pm$0.034 & 0.102$\pm$0.127 & 1.75$\pm$1.48\\
\hline
\end{tabular}
\label{tab1}
\end{center}
\end{table}

\subsubsection{Visual Analysis}
\begin{figure*}
\centering
\includegraphics[width=5.5in, height=3in]{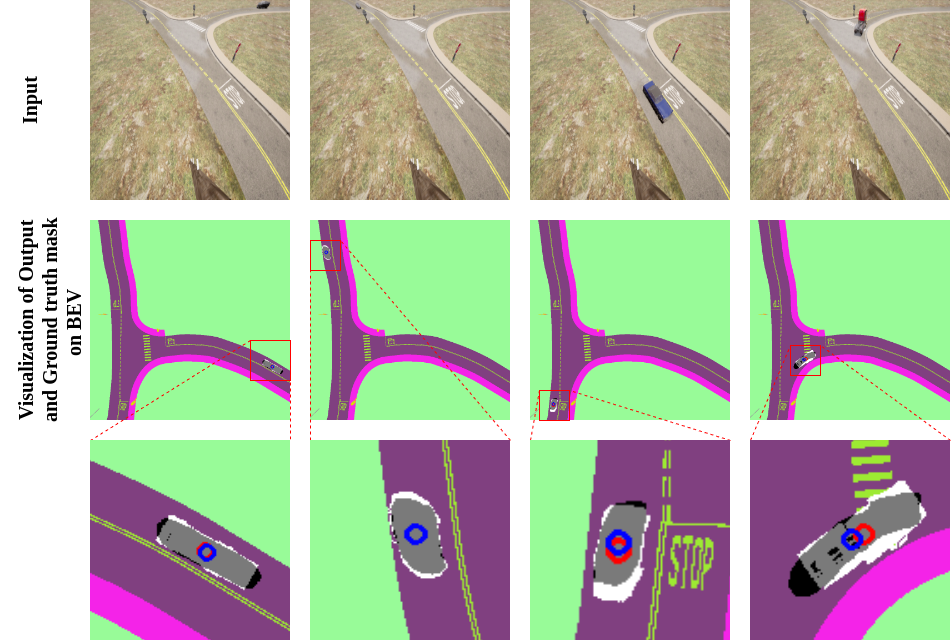}
\caption{Comparison of ground truth mask and predicted mask from the proposed model; the white region in the top view is the ground truth and the black region is the predicted mask from the model. The red circle in the third row represents the centroid of the ground truth mask and the blue circle represents the centroid of the predicted mask.}
\label{fig_comp}
\end{figure*}

For the visual analysis, we overlaid the SDD-UNet's predicted mask and ground truth mask on the segmented BEV image, that was generated from a Carla camera positioned above the intersection. The analysis for various vehicles at different positions is depicted in Fig. \ref{fig_comp}. In this visualization, the ground truth mask is represented in white, while the predicted mask is shown in black, with their overlap appearing in gray. Additionally, centroids are marked with circles on the same image: red circles indicate the centroids of the ground truth masks, and blue circles denote the centroids of the predicted masks. Fig. \ref{fig_comp} clearly demonstrates that the model accurately predicts the perspective transformation of vehicle positions in the BEV. Furthermore, the proximity of the centroids to the ground truth suggests high model accuracy.

\subsubsection{Computational Complexity}
For the hardware, an NVIDIA RTX A2000 GPU is utilized. The model under consideration has a total of 25.6 million parameters and a parameter size of 97.64 MB. The execution time to produce an output is 0.079 seconds. Given these specifications, the model is computationally efficient and well-suited for real-time deployment.

\section{Conclusion}
In this work, we developed a spatial transformer-based model to transform the vision camera perspective to monitor road intersections. Our proposed SDD-UNet model effectively transforms vehicles captured from a static camera in real-world scenarios to a BEV. The SDD-UNet eliminates the distortion faced by the algorithmic approach of perspective transformation in BEV. The SDD-UNet model achieves an average DSC of over 95\%, outperforming the baseline UNet model and the modified UNet with ST as skip connection by 40\% and 10\%, respectively. The centroid of the predicted mask is displaced by only 0.14 meters compared to the ground truth centroid and has the lowest MAE of 0.102 indicating high accuracy. SDD-UNet utilizes a single camera to monitor the junction, a practical setup that is more feasible and cost-effective than a multi-camera system. Thus, the SDD-UNet can cover the gap between BEV simulators and the real-life images from the traffic intersection. Future work could extend this approach to handle multiple vehicles and objects, enhancing position transformation and tracking capabilities.


\bibliographystyle{ieeetr}
\bibliography{refs}

\begin{thebibliography}{10}

\bibitem{icais1}
S.~P. Sahu, D.~K. Dewangan, A.~Agrawal, and T.~Sai~Priyanka, ``Traffic light cycle control using deep reinforcement technique,'' in {\em 2021 International Conference on Artificial Intelligence and Smart Systems (ICAIS)}, pp.~697--702, 2021.

\bibitem{liu1}
X.-Y. Liu, M.~Zhu, S.~Borst, and A.~Walid, ``Deep reinforcement learning for traffic light control in intelligent transportation systems,'' {\em arXiv preprint arXiv:2302.03669}, 2023.

\bibitem{iv}
M.~Klimke, B.~Völz, and M.~Buchholz, ``Automatic intersection management in mixed traffic using reinforcement learning and graph neural networks,'' in {\em 2023 IEEE Intelligent Vehicles Symposium (IV)}, pp.~1--8, 2023.

\bibitem{Reiher2020ASD}
L.~Reiher, B.~Lampe, and L.~Eckstein, ``A sim2real deep learning approach for the transformation of images from multiple vehicle-mounted cameras to a semantically segmented image in bird’s eye view,'' {\em 2020 IEEE 23rd International Conference on Intelligent Transportation Systems (ITSC)}, pp.~1--7, 2020.

\bibitem{sumo}
D.~Krajzewicz, G.~Hertkorn, C.~Feld, and P.~Wagner, ``Sumo (simulation of urban mobility); an open-source traffic simulation,'' pp.~183--187, 01 2002.

\bibitem{highDdataset}
R.~Krajewski, J.~Bock, L.~Kloeker, and L.~Eckstein, ``The highd dataset: A drone dataset of naturalistic vehicle trajectories on german highways for validation of highly automated driving systems,'' in {\em 2018 21st International Conference on Intelligent Transportation Systems (ITSC)}, pp.~2118--2125, 2018.

\bibitem{cvpr1}
L.~Peng, Z.~Chen, Z.~Fu, P.~Liang, and E.~Cheng, ``Bevsegformer: Bird's eye view semantic segmentation from arbitrary camera rigs,'' in {\em Proceedings of the IEEE/CVF Winter Conference on Applications of Computer Vision (WACV)}, pp.~5935--5943, January 2023.

\bibitem{cvpr2}
C.~Pan, Y.~He, J.~Peng, Q.~Zhang, W.~Sui, and Z.~Zhang, ``Baeformer: Bi-directional and early interaction transformers for bird's eye view semantic segmentation,'' in {\em Proceedings of the IEEE/CVF Conference on Computer Vision and Pattern Recognition (CVPR)}, pp.~9590--9599, June 2023.

\bibitem{cvpr3}
R.~Wang, J.~Qin, K.~Li, Y.~Li, D.~Cao, and J.~Xu, ``Bev-lanedet: An efficient 3d lane detection based on virtual camera via key-points,'' in {\em Proceedings of the IEEE/CVF Conference on Computer Vision and Pattern Recognition (CVPR)}, pp.~1002--1011, June 2023.

\bibitem{eccv1}
Z.~Li, W.~Wang, H.~Li, E.~Xie, C.~Sima, T.~Lu, Y.~Qiao, and J.~Dai, ``Bevformer: Learning bird's-eye-view representation from multi-camera images via spatiotemporal transformers,'' in {\em Computer Vision -- ECCV 2022} (S.~Avidan, G.~Brostow, M.~Ciss{\'e}, G.~M. Farinella, and T.~Hassner, eds.), (Cham), pp.~1--18, Springer Nature Switzerland, 2022.

\bibitem{multi_cam}
X.~Huang, P.~He, A.~Rangarajan, and S.~Ranka, ``Machine-learning-based real-time multi-camera vehicle tracking and travel-time estimation,'' {\em Journal of Imaging}, vol.~8, no.~4, 2022.

\bibitem{lidar1}
B.~Cherif, H.~Ghazzai, A.~Alsharoa, H.~Besbes, and Y.~Massoud, ``Aerial lidar-based 3d object detection and tracking for traffic monitoring,'' in {\em 2023 IEEE International Symposium on Circuits and Systems (ISCAS)}, pp.~1--5, 2023.

\bibitem{2023arXiv230301212S}
Y.~{Shi}, K.~{Jiang}, J.~{Li}, J.~{Wen}, Z.~{Qian}, M.~{Yang}, K.~{Wang}, and D.~{Yang}, ``{Grid-Centric Traffic Scenario Perception for Autonomous Driving: A Comprehensive Review},'' {\em arXiv e-prints}, p.~arXiv:2303.01212, Mar. 2023.

\bibitem{lidar2}
J.~Zhao, H.~Xu, Y.~Tian, and H.~Liu, ``Towards application of light detection and ranging sensor to traffic detection: an investigation of its built-in features and installation techniques,'' {\em Journal of Intelligent Transportation Systems}, vol.~26, no.~2, pp.~213--234, 2022.

\bibitem{OpenStreetMap}
{OpenStreetMap contributors}, ``{Planet dump retrieved from https://planet.osm.org }.'' \url{ https://www.openstreetmap.org }, 2017.

\bibitem{roadrunner}
J.~Jeong, N.~Kim, D.~Karbowski, and A.~Rousseau, ``Implementation of model predictive control into closed-loop micro-traffic simulation for connected automated vehicle,'' {\em IFAC-PapersOnLine}, vol.~52, no.~5, pp.~224--230, 2019.
\newblock 9th IFAC Symposium on Advances in Automotive Control AAC 2019.

\bibitem{carla}
A.~Dosovitskiy, G.~Ros, F.~Codevilla, A.~Lopez, and V.~Koltun, ``{CARLA}: {An} open urban driving simulator,'' in {\em Proceedings of the 1st Annual Conference on Robot Learning}, pp.~1--16, 2017.

\bibitem{googlemaps}
{Google Maps}, ``Google maps.'' https://maps.app.goo.gl/iqaKfq3oKzWj2Tom8.
\newblock Accessed: 2024-04-28.

\bibitem{unet}
O.~Ronneberger, P.~Fischer, and T.~Brox, ``U-net: Convolutional networks for biomedical image segmentation,'' in {\em Medical image computing and computer-assisted intervention--MICCAI 2015: 18th international conference, Munich, Germany, October 5-9, 2015, proceedings, part III 18}, pp.~234--241, Springer, 2015.

\bibitem{yolo}
G.~Jocher, A.~Chaurasia, and J.~Qiu, ``{Ultralytics YOLO},'' Jan. 2023.

\bibitem{ST}
M.~Jaderberg, K.~Simonyan, A.~Zisserman, {\em et~al.}, ``Spatial transformer networks,'' {\em Advances in neural information processing systems}, vol.~28, 2015.

\bibitem{opencv_library}
G.~Bradski, ``The opencv library,'' {\em Dr. Dobb's Journal of Software Tools}, 2000.

\end{thebibliography}


\end{document}